\def\year{2019}\relax
\documentclass[letterpaper]{article} 
\usepackage{aaai19}  
\usepackage{times}  
\usepackage{helvet}  
\usepackage{courier}  
\usepackage{url}  
\usepackage{graphicx}  

\usepackage{booktabs}
\usepackage{amsfonts}
\usepackage[table,xcdraw]{xcolor}
\usepackage{multirow}

\usepackage{amsmath}
\usepackage{makecell}

\usepackage[colorlinks,linkcolor=black,anchorcolor=black,citecolor=black,urlcolor=blue]{hyperref}

\begin{document}
%
\title{Domain-Adversarial Multi-Task Framework for \\ Novel Therapeutic Property Prediction of Compounds}
\author{Lingwei Xie,$^{\textbf{1,\dag,*}}$ Song He,$^\textbf{2,\dag}$ Shu Yang,$^\textbf{3}$ Boyuan Feng,$^\textbf{3}$ Kun Wan,$^\textbf{3}$ Zhongnan Zhang,$^{\textbf{1,*}}$ Xiaochen Bo,$^{\textbf{2,*}}$ Yufei Ding$^\textbf{3}$\\
$^1$Software School, Xiamen University, Xiamen, China, 361005\\
$^2$Beijing Institute of Radiation Medicine, Beijing, China, 100850\\
$^3$Computer Science, University of California, Santa Barbara, California, 93106\\
xielingwei@stu.xmu.edu.cn, zhongnan\_zhang@xmu.edu.cn \\
809848790@qq.com, boxiao@163.com\\
\{shuyang1995, boyuan, kun, yufeiding\}@cs.ucsb.edu
}
\maketitle
\begin{abstract}
With the rapid development of high-throughput technologies, parallel acquisition of large-scale drug-informatics data provides huge opportunities to improve pharmaceutical research and development. One significant application is the purpose prediction of small molecule compounds, aiming to specify therapeutic properties of extensive purpose-unknown compounds and to repurpose novel therapeutic properties of FDA-approved drugs. Such problem is very challenging since compound attributes contain heterogeneous data with various feature patterns such as drug fingerprint, drug physicochemical property, drug perturbation gene expression. Moreover, there is complex nonlinear dependency among heterogeneous data. In this paper, we propose a novel domain-adversarial multi-task framework for integrating shared knowledge from multiple domains. The framework utilizes the adversarial strategy to effectively learn target representations and models their nonlinear dependency. Experiments on two real-world datasets illustrate that the performance of our approach obtains an obvious improvement over competitive baselines. The novel therapeutic properties of purpose-unknown compounds we predicted are mostly reported or brought to the clinics. Furthermore, our framework can integrate various attributes beyond the three domains examined here and can be applied in the industry for screening the purpose of huge amounts of as yet unidentified compounds. Source codes of this paper are available on \href{https://github.com/JohnnyY8/Domain-Adversarial-Multi-task-Framework.git}{Github}.
\end{abstract}

\section{Introduction}
Purpose prediction of small molecule compounds is critical in pharmaceutical research and development (R\&D) \cite{macarron2011impact}. While considerable progress has been achieved during the past decades, traditional de novo strategy remains to be extremely costly, risky, and time consuming \cite{dimasi2003price}, especially for pharmaceutical companies who synthesize and accumulate huge amount of small molecule compound without a definite therapeutic property. To conduct purpose prediction economically attractive, low-risking, and time-saving, drug-discovering strategies based on drug-informatics data and computational approaches have been introduced and exhibits substantial improvement \cite{keiser2009predicting,boguski2009repurposing}.

It is a challenging problem to specify therapeutic properties of purpose-unknown compounds and to repurpose novel therapeutic properties of FDA-approved drugs since each kind of compound contains complicated attributes (See Fig. 1 as an example), such as drug fingerprint, drug physicochemical property, drug perturbation gene expression and so on. Many researchers tried only chemical structure for prediction \cite{ma2013drug,haupt2011old,xie2017discovery,wang2016drug}. Nevertheless, Yildirim \textit{et al}. pointed out that most drugs with the same targets have different chemical structures so that structure-based prediction seems not so convincing \cite{yildirim2007drug}. Moreover, the constitution of each individual attribute varies significantly. For example, gene expression consists of over 10,000 genome-wide expressions while the structural fingerprint is a kind of graphical representation of arrangement of chemical bonds between atoms. Besides, there is high level of dependency among different attributes. For instance, drug fingerprint not only refers to the inside spatial arrangement of atoms, also leads to various physical and chemical properties. It turns out that these complicated drug attributes yield heterogeneous drug data consisting of correlated domains, and data patterns of which vary significantly.

\begin{figure}
  \centering
  	\includegraphics[width=0.45\textwidth]{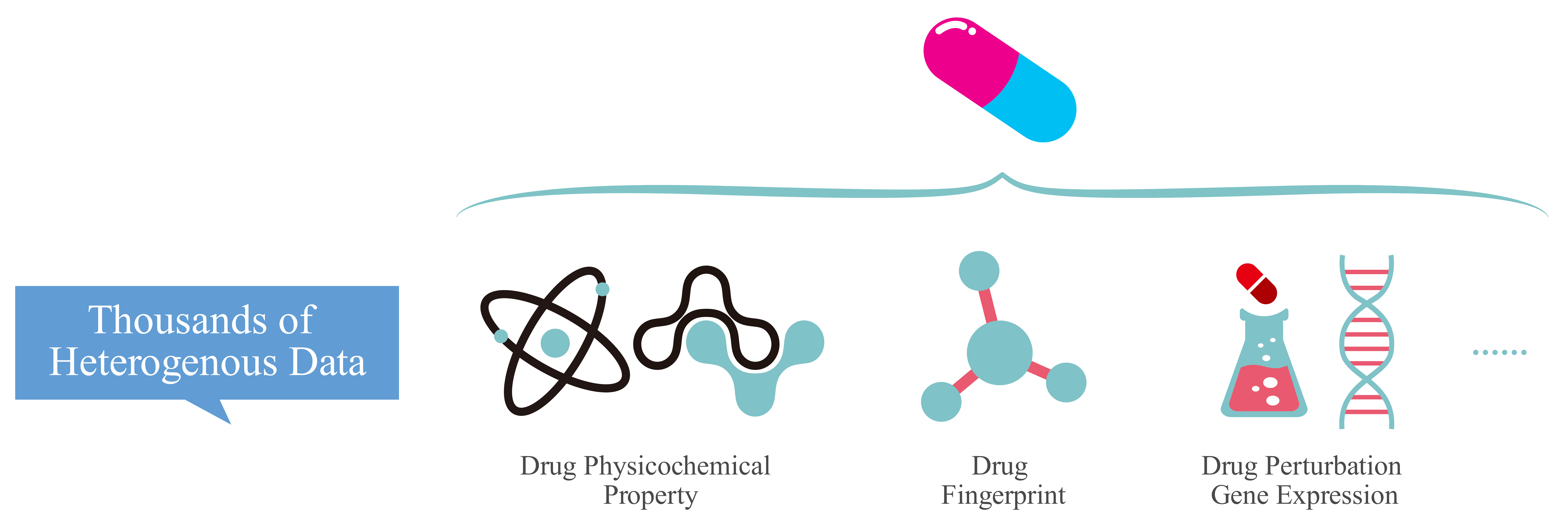}
  	\caption{Heterogeneous drug-informatics data. The attributes of drugs vary significantly from each other, but underlying complementary information exists among them.}
\end{figure}

Most existing approaches for integrating drug-informatics data from several domains were based on the linear combination. For example, Napolitano \textit{et al}. constructed three drug similarity networks based on drug structure, distance of drug targets in protein-protein interaction networks, and expression patterns of drug perturbations, then integrated datasets by averaging three drug similarity measurements to predict new therapeutic properties of drugs \cite{napolitano2013drug}. Wang \textit{et al}. proposed a new algorithm, named PreDR, which predicts unidentified drug-disease associations by taking the maximums of three drug similarity matrices derived from chemical structure, target protein sequence, and side effect profile similarities \cite{wang2013drug}. However, the performance of these approaches reach a plateau because of inefficient compatibility. Although some methods have improvements by integrating two data domains, they are barely generalized to others, and the performance even degrades when scale to more data domains. On the one hand, the intrinsic nonlinear patterns which possess better prospect for inference would be ignored inevitably when linear combination. On the other hand, there are many overlap information among data domains.

In the literature, joint learning representations of heterogeneous data is widely studied especially for speech recognition and clinical endpoint prediction. The state-of-the-art approaches are commonly extensions of recurrent neural networks (RNNs) with the Long Short-term Memory (LSTM) units. RNNs are commonly used for modeling many kinds of temporal sequences and the correlations among multiple data. It is nontrivial to apply to data without time attribute, and the irregular data shapes are still a challenge for RNNs. Motivated by advances in training deep neural networks (DNNs) and the availability of large datasets, we seek an approach that is able to: (1) effectively model the irregular heterogeneous data; (2) parse the underlying complex nonlinear dependency among heterogeneous data; (3) joint learning and scale up numbers of different data domains.

Currently, some efforts have been made to embed heterogeneous data into the same feature space, but the embedding layers are usually used to extract the domain-invariant features. Inspired by the success of adversarial strategy on domain adaption \cite{ajakan2014domain,ganin2016domain,bousmalis2016domain}, we use adversarial strategy to make sure the shared layers extract domain-specific features which are suitable for each data domain. Specifically, we propose a domain-adversarial extractor and regard domain-adversarial learning as a classification task. Finally, each type of data are embedded with domain-specific features and fed into next stage, then the dependency among heterogeneous data is modeled for knowledge sharing. Compared to the vanilla convolutional neural networks (CNNs) or hybrid models that combine CNNs and RNNs, our proposed framework extends it by joint learning the representations of heterogeneous data under a domain-adversarial multi-task framework.

We conduct experiments on heterogeneous data of many FDA-approved drugs and purpose-unknown compounds. The results on five evaluation metrics (Hamming Loss, One Error, Coverage, Ranking Loss, and Average Precision) prove that our method holds the great promise to identity the purpose of small molecule compounds more accurately. Moreover, we predict novel therapeutic properties for more than 3,000 unidentified compounds, and the prediction results are partly reported by other researches. As a highly extensible method, our framework can integrate various data beyond the three domains examined here.

In the paper, our main contributions are:
\begin{itemize}
\item We approach the purpose prediction of small molecule compounds problem as a representation learning task based on drug-informatics data that are from different domains. To our knowledge, this is the first time to predict novel therapeutic properties of both extensive purpose-unknown compounds and FDA-approved drugs by multi-task learning with adversarial strategy.
\item We design a domain-adversarial extractor for learning domain-specific features of heterogeneous data. Then our proposed framework joint learns representation for integrating multiple domain knowledge. Eventually, five evaluation metrics are used to assess the performance of proposed framework in its entirety.
\item We conduct experiments of real-world data (including drug physiochemical property, drug fingerprint, and drug perturbation gene expression) on the task of novel therapeutic property prediction. By the conclusion of training on the real-world existing date, promising results prove the effectiveness of our proposed framework over competitive baselines, and novel properties of compounds we predicted are mostly in line with their patents.
\end{itemize}

\begin{table}[h]
\caption{Mathematical Notations}
\begin{tabular}{ cc }
\arrayrulecolor{black}\hline
\rowcolor[HTML]{9B9B9B}
{\color[HTML]{FFFFFF} {\textbf{Symbol}}} & {\color[HTML]{FFFFFF}{\textbf{Description}}} \\

\arrayrulecolor{black}
\hline
\begin{tabular}[c]{ p{2.8cm} }  \textit{X: $x^i\in$X, \#\{X\}=m}; \\ \textit{Y: $y^i\in$Y, $y^i\in$ $\mathbb{R}^{q}$};  \end{tabular} &
\begin{tabular}[c]{ p{4.3cm} }  {indices and number of data;}\\ {indices and number of classes;}  \end{tabular} \\

\rowcolor[HTML]{C0C0C0}
\begin{tabular}[c]{ p{2.8cm} }\rowcolor[HTML]{C0C0C0} \\ \textit{F: $f^i\in$F}, \\ \textit{$f^i$ = \{$f^i_0$, $f^i_1$,.., $f^i_k$ \}}, \\ \textit{$f^i_j$ = (Value, Type)}; \\ \\ \end{tabular} & \begin{tabular}[c]{ p{4.3cm} }feature space,\\ each datum has a feature set,\\ each feature element consists of value and type;\end{tabular} \\

\begin{tabular}[c]{ p{2.8cm} } \textit{$e^{x^i}$ = \{$e^{f^i_j} | j\in$k\}}; \end{tabular} & \begin{tabular}[c]{ p{4.3cm} } embedding vectors; \end{tabular} \\

\rowcolor[HTML]{C0C0C0}
\begin{tabular}[c]{ p{2.8cm} }\rowcolor[HTML]{C0C0C0} \\ \textit{$E(f^i)$=\{$E(f^i_j) | j\in$k\}}, \textit{$L_{Ext}(\theta_{Ext},X)$}; \\ \\ \end{tabular} & \begin{tabular}[c]{ p{4.3cm} } \\ functions for domain-specific extractor; \\ \\ \end{tabular} \\

\begin{tabular}[c]{ p{2.8cm} } \textit{C($x^i$) = C($f^i$)}, \\ \textit{$L_{Cls}(\theta_{Cls},X)$}; \\ \end{tabular} & \begin{tabular}[c]{ p{4.3cm} } functions for classification; \\ \end{tabular} \\
\arrayrulecolor{black}\hline
\end{tabular}
\end{table}

\section{Problem Definition}
In this research, we integrate heterogeneous data, which means each data corresponds to several kinds of features, to predict therapeutic properties of purpose-unknown compounds and repurpose novel therapeutic properties of FDA-approved drugs. The whole framework consists of two stages. Learning target representation for each kind of feature in the first stage is modeled as a multi-class classification task when each feature type is encoded as one-hot vectors. In the second stage, therapeutic property prediction for each data is regarded as one multi-label classification task according to known uses of small molecule compounds, and the number of known uses can be at least one. Some symbols and descriptions of tasks as in Table 1.

\section{Proposed Method}
In this section, we discuss the whole framework (as shown in Fig. 2), including the heterogeneous data embedding, domain-adversarial extractor for more specific domain information, and joint learning under multi-task framework.

\begin{figure}[h]
  \centering
  	\includegraphics[width=0.45\textwidth]{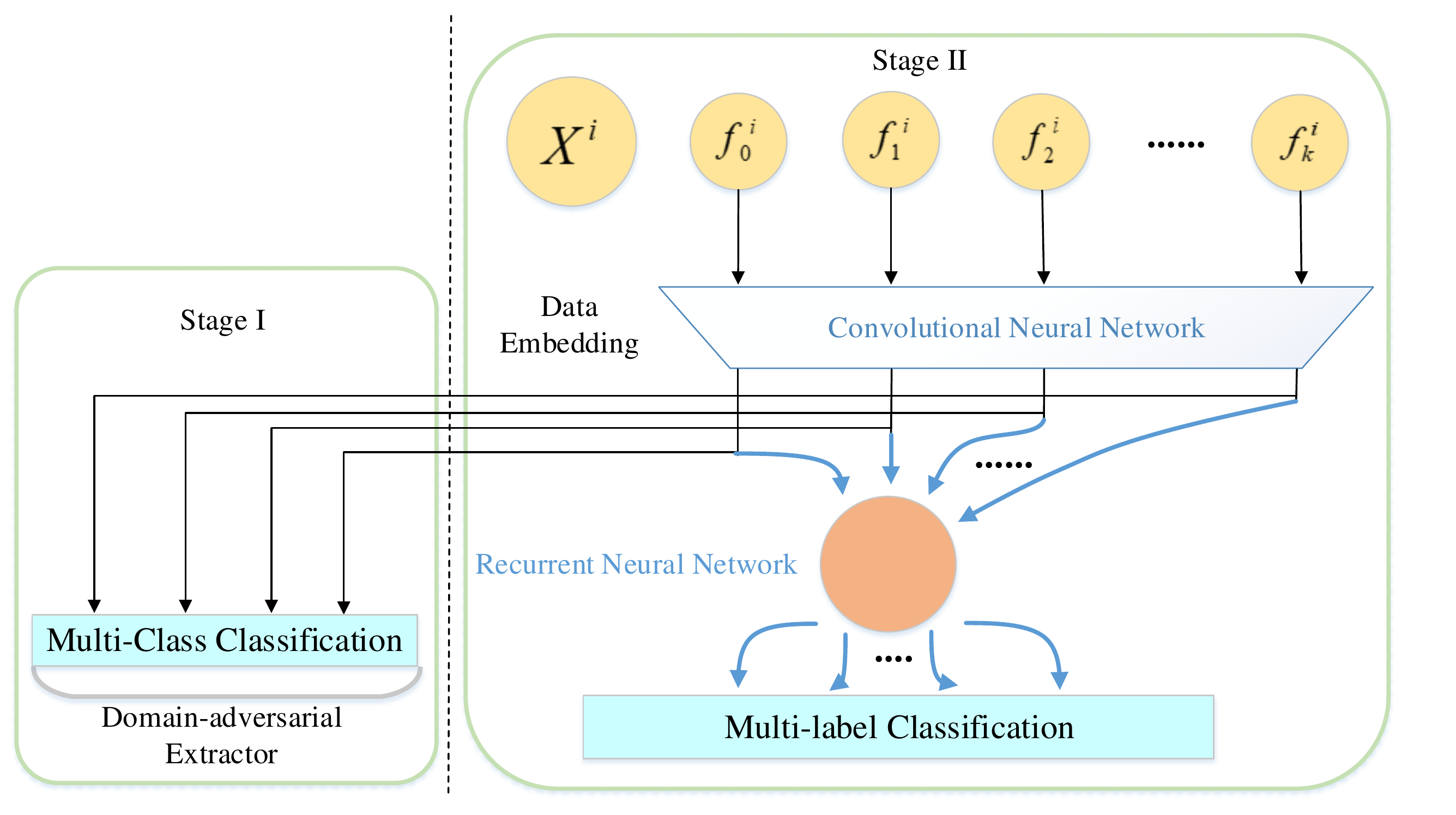}
  	\caption{The training for the whole framework consists of two stages. Stage I is a multi-class classification task for extracting domain-specific features, and Stage II is a multi-label classification task for integrating knowledge of multiple domains under the multi-task framework.}
\end{figure}

First of all, fix-sized feature vectors are generated from different domains by embedding, during which a domain-adversarial extractor is utilized to extract more domain-specific information. Then we integrate shared knowledge from multiple domains by regarding each class as a single binary classification task. Accordingly, we combine stages as an ensemble, as defined in Equation (1), to address the challenge of novel therapeutic property prediction of small molecule compounds under a domain-adversarial multi-task framework.

\begin{equation}
\underset{\theta}{\operatorname{argmin}} \textit{\textbf{L}$(\theta;X)=\textbf{L}_{Ext}(\theta_{Ext},X)+\textbf{L}_{Cls}(\theta_{Cls},X)$}
\end{equation}

\subsection{Heterogeneous Data Embedding}
CNNs are well-known deep learning architecture for encoding input with arbitrary size in a sliding-window manner, and the DeepSEA successfully applied CNNs to sequence-based problems in genomics \cite{zhou2015predicting}. There are 6 layers before classification in the entire architecture, including 3 convolutional layers (320, 480, and 960 convolutional kernels individually), 2 max pooling layers, and 1 fully connected (FC) layer. DeepSEA predicts chromatin features after training with large-scale chromatin-profiling data from the ENCODE project \cite{encode2012integrated}.

\begin{figure}[h]
  \centering
  	\includegraphics[width=0.45\textwidth, height = 9cm]{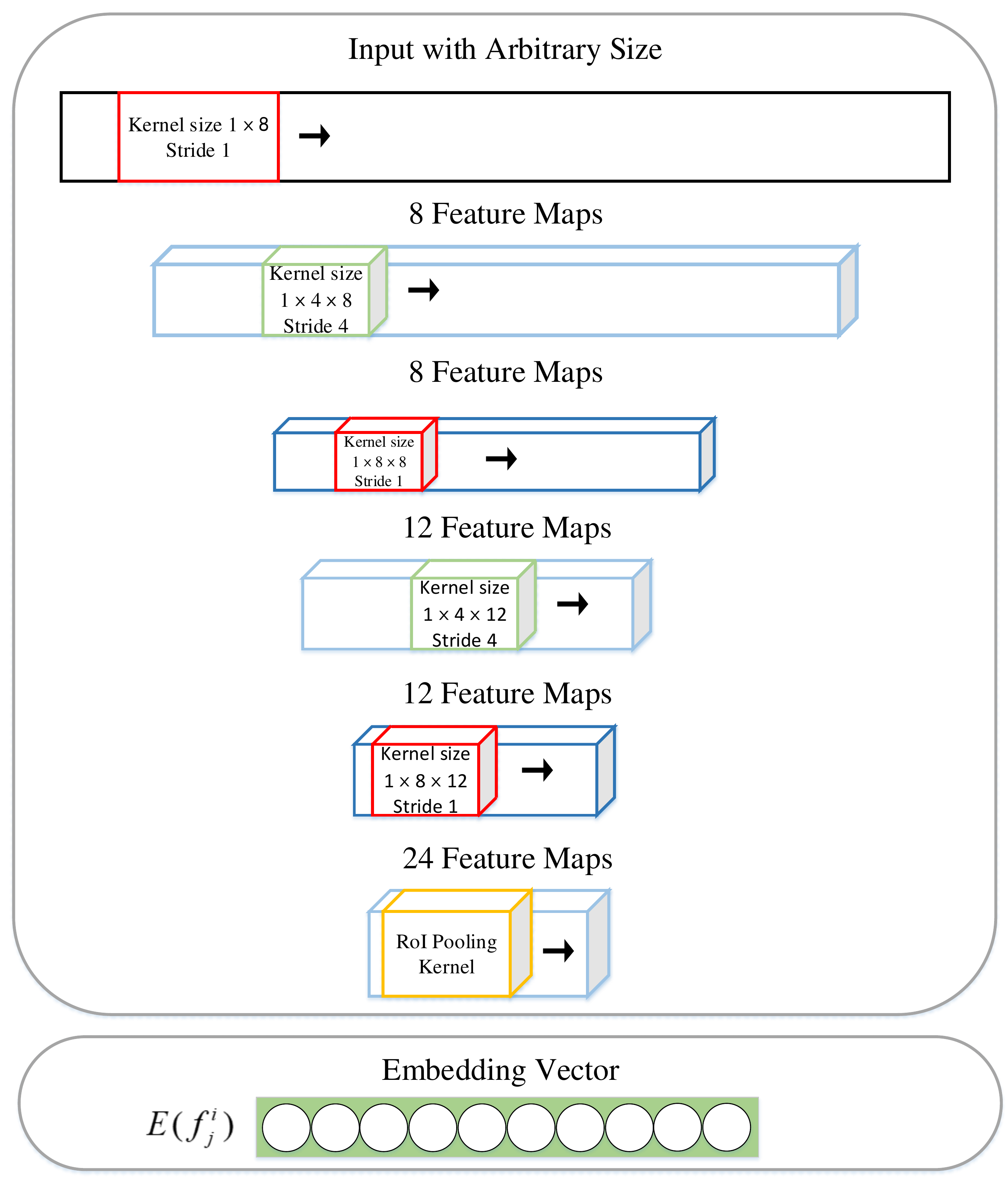}
  	\caption{CNN architecture. A CNN mainly consists of three parts: convolutional layers, max pooling layers, and fully connected layers. In this paper, we use RoI pooling layer instead of fully connected layers to extract fixed-length feature vector from the feature maps.}
\end{figure}

Owing to its outstanding performance on genomic sequences, we simplify original architecture for embedding heterogeneous data that have various feature patterns. As shown in Fig. 3, the basic feature extractor of CNN are stacked by convolution operations with different number of kernels and max pooling layers, then the fully connected layer has been replaced by region of interest (RoI) pooling layer to generate fixed dimensional embedding vectors \cite{girshick2015fast}. On the one hand, the capacity of learning more abstract patterns could be guaranteed by each convolutional layer, and the receptive field of top layer should be smaller than original sequence. On the other hand, the number of parameters from full connected layers accounts for about 80\% in whole neural network, they will enlarge the model size and slow down the model training. So both reduction for number of convolutional kernels and RoI pooling layer as substitution bring several benefits, especially decreasing overfitting risk.

\subsection{Domain-Adversarial Extractor}
To learn target representations, we assume that embedding vectors are classified accurately about which type they belong to if they consist of more differences. Because in our area, data are heterogeneous with different feature patterns and describing different attributes, due to various data sources. For instance, drug physiochemical property are from JoeLib, OpenBabel and Chemminer chemoinformatics databases; drug fingerprints are from PubChem Compound database; and drug perturbation gene expression are from L1000 database of LINCS project. Each type of feature contains similar and different characters. Although heterogeneous data are transformed into the same feature space after embedding, there is commonly no guarantee to keep more domain-specific information.

\begin{equation}
p(\cdot|e^{f^i_j})=Softmax(\textbf{\textit{W}}^{\rm{T}}\textbf{\textit{h}$_X$}+\textbf{\textit{b}})
\end{equation}

\begin{figure}[h]
  \centering
  	\includegraphics[width=0.45\textwidth]{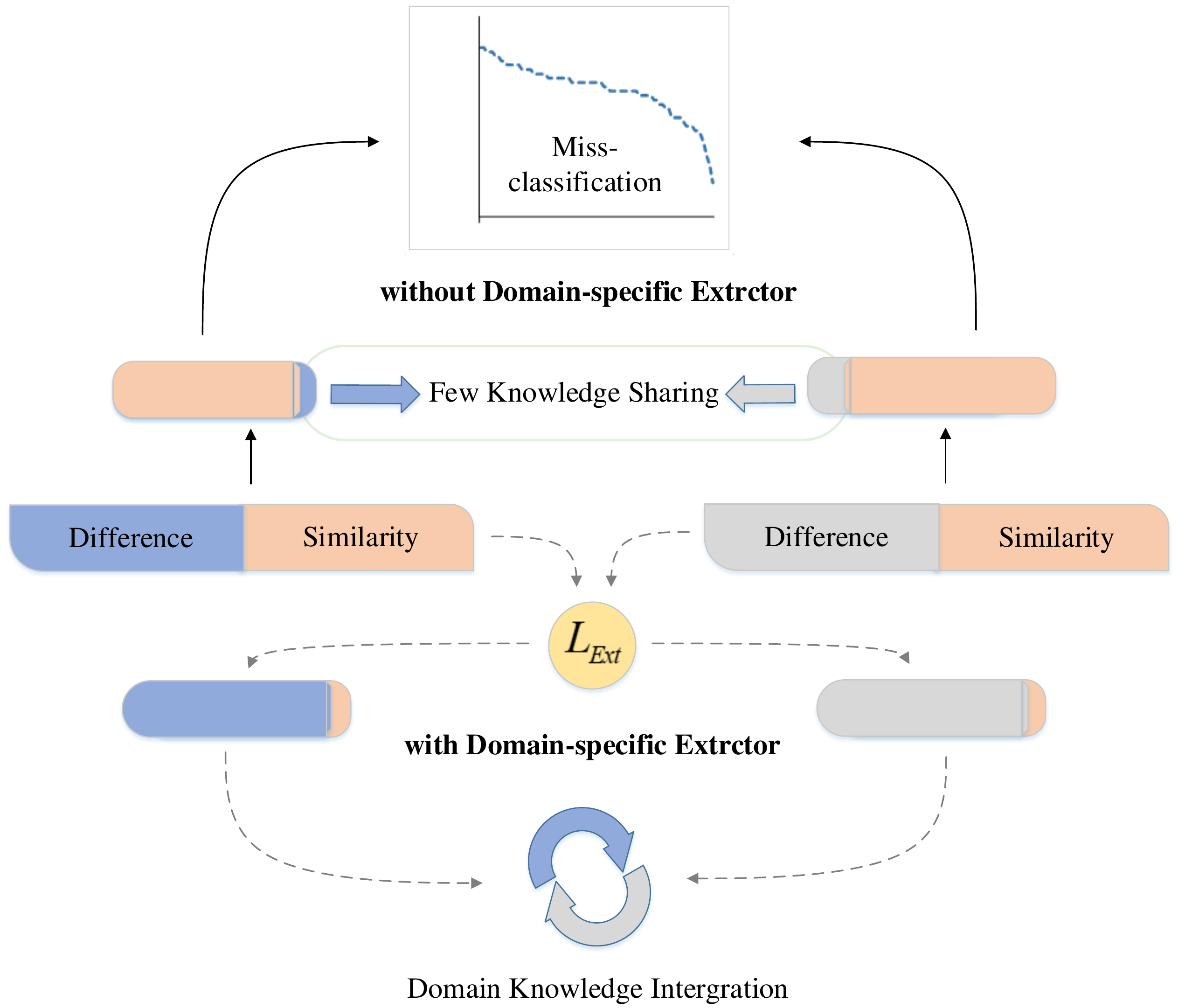}
  	\caption{Domain-adversarial learning. The inefficient feature extraction leads to too much similar information in vectorial representations for many existing approaches. But after domain-specific extractor gets fully trained, the domain-invariant redundancy is reduced effectively.}
\end{figure}

Inspired by every single type of data is important in its own way and provides different views of multiple domains, we devise a domain-adversarial extractor to make heterogeneous data provide domain-specific instead of domain-invariant information for integration. During domain-adversarial learning, each type of feature is extracted with more individual private characters that represent special domain knowledge. Otherwise, the shared layers pay more attention to similarities which introduce redundancy for embedding (as shown in Fig. 4). Specifically, given an embedding vector $e^{f^i_j}$, we use a Softmax model which introduces adversarial competition among classes to compute the probability $p(\cdot|e^{f^i_j})$ over all features, as defined in Equation (2). After domain-adversarial extractor gets fully trained, the pre-trained model is regarded as prior knowledge for learning in the next stage.

\subsection{Joint Learning under Multi-task Framework}
As an extension of RNNs, bidirectional LSTM (Bi-LSTM) is a strong and trainable sequence model. In this paper, it is a special case for Bi-LSTM to integrate all heterogeneous data by regarding each of them as an input in one time step, because there is no temporal extension for all features. At meantime, the dependency among different types of features is parsed over the positive time and negative time direction, as defined in Equation (3). Finally, Bi-LSTM incorporates information and transfers embedding vectors into high-level representations by concatenating two final states, as shown in Fig. 5.

\begin{equation}
\begin{aligned}
\textbf{h}_t &= \overrightarrow{\textbf{h}}_t \oplus \overleftarrow{\textbf{h}}_t \\
&= \textbf{Bi-LSTM}(e^{x^i}, \overrightarrow{\textbf{h}}_{t-1}, \overleftarrow{\textbf{h}}_{t+1}, \theta_{Cls})
\end{aligned}
\end{equation}

where $\overrightarrow{\textbf{h}}_t$ and $\overleftarrow{\textbf{h}}_t$ are the hidden states at time \textit{t} of the forward and backward LSTMs respectively; $\oplus$ is concatenation operation; $\theta$ denotes all parameters in Bi-LSTM.

\begin{figure}[h]
  \centering
  	\includegraphics[width=0.45\textwidth]{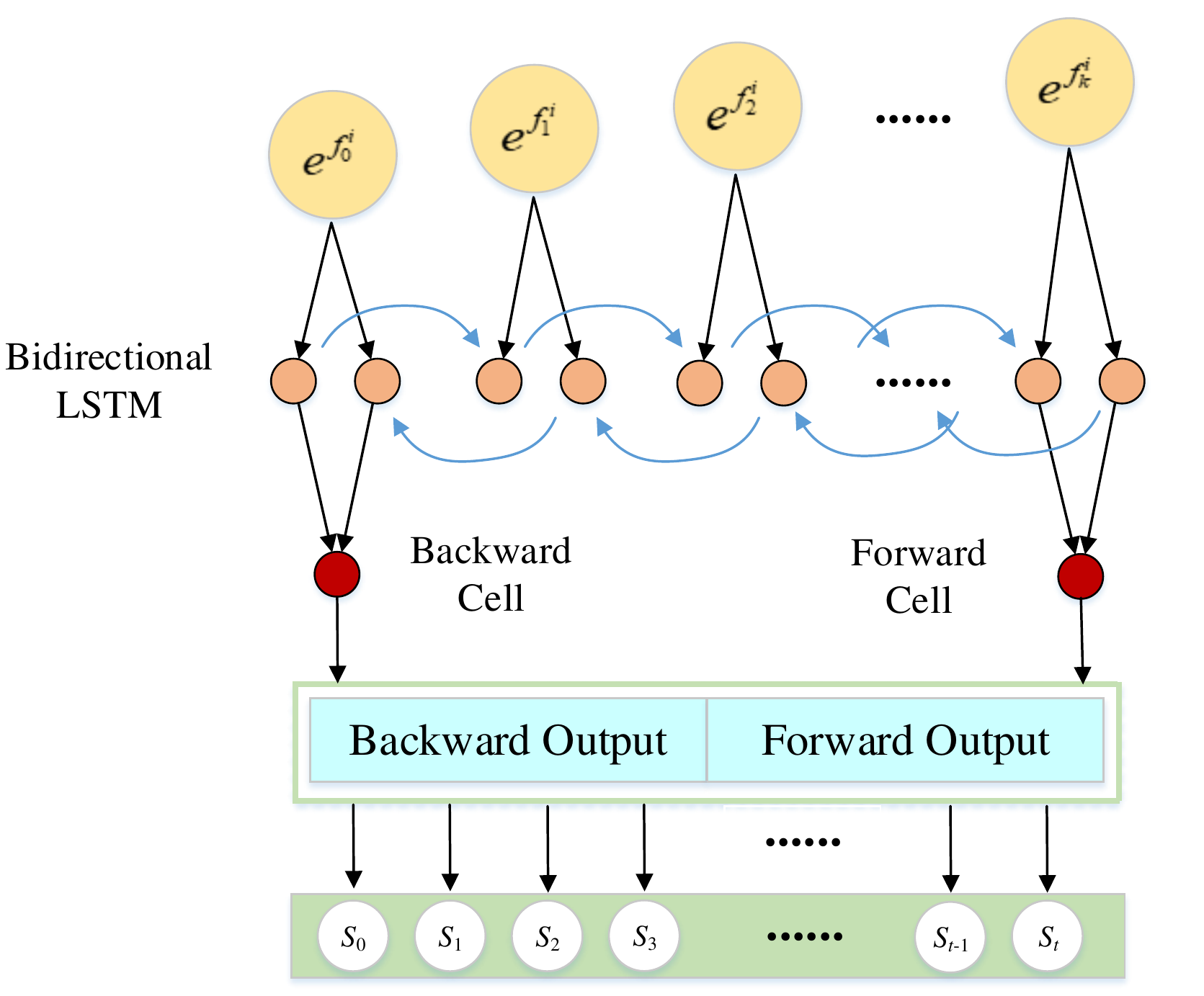}
  	\caption{Bi-LSTM architecture. Bi-LSTM can be trained using all available information in both forward and backward direction. Last, binary classifiers share knowledge with each other to make model joint learn representations under the multi-task framework.}
\end{figure}

\begin{equation}
\begin{aligned}
\textbf{\textit{L}}_{Cls}(\theta_{Cls},X) = & \max(C(X), 0)-C(X) \times Y \\
&+ \log(1 + \exp(-\mid C(X) \mid))
\end{aligned}
\end{equation}

To learn joint representations of heterogeneous data, we build a multi-task framework for all classes. Specifically, target prediction is formulated as a multi-label classification problem, and each of class is regarded as related binary classification simultaneously under the multi-task framework. Because the information coming from the training signals of related tasks are inevitably ignored by ensembling separated models. Then the Bi-LSTM is trained with objective function of sigmoid cross entropy, as defined in Equation (4), resulting in effective knowledge sharing under the multi-task framework for novel therapeutic property prediction of compounds.

\section{Experiments}
\subsection{Data Description}
We set up two datasets for evaluating our model. The first one contains three domain-specific FDA-approved drugs (\textbf{TDF}) of 6,025 samples representing 637 FDA-approved drugs. Another one contains three domain-specific unidentified compounds (\textbf{TDU}) of 3,749 samples representing 3,749 unidentified compounds. Three domain-specific data record drug perturbation gene expression level, structural fingerprint, and physicochemical property separately. The details as following:

\textbf{Drug Perturbation Gene Expression}: from L1000 database of LINCS project, we extract Z-scores of gene expression data when FDA-approved drugs and as yet purpose-unknown compounds are exposed in five typical cell lines (A375, HA1E, HT29, MCF7, and PC3). In TDF, we regard gene expression data stimulated by compounds with different doses and times as different samples, amounting to 6,025 transcriptional expression samples for each cell line. In TDU, we combined gene expression data stimulated by different dose of compounds by a weighted average way introduced by Subramanian A \textit{et al} \cite{subramanian2017next}.

\textbf{Drug Fingerprint}: the structure data in SDF format are from PubChem Compound database and used to calculate the fingerprint vector of each drug \cite{kim2015pubchem,cao2008chemminer}. Features in this domain represent chemical structure of these compounds. In TDF, we replicate the fingerprint vector for the same FDA-approved drugs to match each sample of gene expression domain. In TDU, each fingerprint vector represents a unique purpose-unknown compound.

\textbf{Drug Physicochemical Property}: the JoeLib, OpenBabel and Chemminer chemoinformatics databases provide the information to calculate the physicochemical property and chemical descriptor of compounds. Features in this domain represent the chemical and physical properties, such as the molecular weight, polar surface area, etc. In TDF, we replicate the physicochemical property vector for the same FDA-approved drugs to match each sample of gene expression domain. In TDU, each physicochemical property vector represents a unique purpose-unknown compound.

The first level of drug ATC codes, which indicate drug therapeutic properties, is used to label each drug sample. Notably, each drug has no less than one ATC labels (A: alimentary tract and metabolism; B: blood and blood-forming organs; C: cardiovascular system; D: dermatologicals; G: genito-urinary system and sex hormones; H: systemic hormonal preparations, excluding sex hormones and insulins; J: anti-infectives for systemic use; L: anti-neoplastic and immunomodulating agents; M: musculoskeletal system; N: nervous system; P: anti-parasitic products; R: respiratory system; S: sensory organs; and V: several others). The ATC labels are mostly obtained from the records of DrugBank database \cite{wishart2017drugbank}, and the rest of them are obtained from PubChem Compound database.

\begin{table*}[htbp]
\begin{center}
\caption{Mean Performance Comparisons Across Methods.}
	\begin{tabular}{lllllll}
        \toprule
        &  & Hamming Loss$\downarrow$ & One Error$\downarrow$ & Coverage$\downarrow$ & Ranking Loss$\downarrow$ & Average Precision$\uparrow$ \\
        \toprule
        \multirow{3}{*}{\textbf{RF}} & Gene expression  & 0.3438 & 0.6238 & 3.0347 & 0.2051 & 0.5479 \\ 
                                     & Linear combination & 0.3261 & 0.5809 & 2.7822 & 0.1865 & 0.5786 \\ 
                                     & Physicochemical property & 0.0785  & 0.4307 & 1.8003 & 0.11 & 0.7011 \\ \hline
        \multirow{3}{*}{\textbf{GBDT}} & Gene expression  & 0.1116 & 0.5314 & 3.1914 & 0.2132 & 0.5971 \\ 
                                       & Linear combination & 0.1541 & 0.3581 & 1.9703 & 0.118 & 0.7357 \\ 
                                       & Physicochemical property & 0.0383  & 0.2228 & 1.2558 & 0.0697 & 0.8346 \\
        \hline

        \multirow{2}{*}{\textbf{RF}} & AE & 0.3035 & 0.3812 & 1.5528 & 0.0904 & 0.7363 \\ 
                                     & PCA & 0.3533  & 0.6073 & 2.6634 & 0.1788 & 0.5777 \\ \hline
        \multirow{2}{*}{\textbf{GBDT}} & AE & 0.0539 & 0.1436 & 1.1287 & 0.0588 & 0.8814 \\ 
                                       & PCA & 0.0779  & 0.1419 & 1.0182 & 0.0499 & 0.8865 \\ \hline
        \multicolumn{2}{l}{\textbf{CNN}} & 0.0807  & 0.9356 & 0.9637 & 0.5154 & 0.3342 \\  \hline
        \multicolumn{2}{l}{\textbf{\begin{tabular}[c]{@{}l@{}}CNN + Bi-LSTM\end{tabular}}} & \textbf{0.0153}  & \textbf{0.0809} & \textbf{0.3086} & \textbf{0.0135} & \textbf{0.9499} \\
        \hline

        \multicolumn{2}{l}{\textbf{\begin{tabular}[c]{@{}l@{}}CNN + Bi-LSTM\\ (Domain-specific extractor)\end{tabular}}} & \textbf{0.0083}  & \textbf{0.0479} & \textbf{0.2244} & \textbf{0.0072} & \textbf{0.9710} \\
        \bottomrule
	\end{tabular}
\end{center}
\end{table*}

\subsection{Training and Evaluation Settings}
In this paper, the whole multi-task framework is trained over two stages using Dataset 1. The first stage is to make CNN extract domain-specific features. Thus, we discriminately pre-trained the CNN based on randomly initialized parameters by introducing adversarial competition among several feature types. Once domain-specific features are extracted, the performance of Softmax model nearly matches our assumption that feature types can be classified accurately.

In the second stage, we apply Bi-LSTM to model underlying nonlinear dependency existing among heterogeneous data based on domain-specific feature vectors. Followed by a multi-label classification layer which consists of multiple binary classifiers corresponding to each class. So during the training procedure of Bi-LSTM, the CNN also get further optimization and all classes joint learn representation by sharing the specific domain knowledge. As a result, the prediction performance gets further improvement on several multi-label classification evaluation metrics (as defined in Equation 5$\sim$9) \cite{schapire2000boostexter}.

(1) \textbf{Hamming Loss}: evaluates how many times an instance-label pair is misclassified, i.e. a label not belonging to the instance is predicted or a label belonging to the instance is not predicted. The performance is perfect when hamming loss equals to 0; the smaller the value of hamming loss, the better the performance.

\begin{equation}
\begin{aligned}
HammingLoss=\frac{1}{m} \sum_{i=1}^{m} \frac{1}{q} \left| C(x^i) \bigtriangleup y^i \right|
\end{aligned}
\end{equation}

(2) \textbf{One Error}: evaluates how many times the top-ranked label is not in the set of proper labels of the instance. The performance is perfect when one error equals to 0; the smaller the value of one error, the better the performance.

\begin{equation}
\begin{aligned}
OneError=\frac{1}{m} \sum_{i=1}^{m} [\![ [\underset{\hat{y} \in Y}{\operatorname{argmax}}C(x^i,\hat{y})] \notin y^i]\!]
\end{aligned}
\end{equation}

(3) \textbf{Coverage}: evaluates how far we need, on the average, to go down the list of labels in order to cover all the proper labels of the instance. It is loosely related to precision at the level of perfect recall. The smaller the value of coverage, the better the performance.

\begin{equation}
\begin{aligned}
Coverage=\frac{1}{m} \sum_{i=1}^{m} \underset{\hat{y} \in y^i}{\max} rank[C(x^i, \hat{y})] - 1
\end{aligned}
\end{equation}

(4) \textbf{Ranking Loss}: evaluates the average fraction of label pairs that are reversely ordered for the instance. The performance is perfect when ranking loss equals to 0; the smaller the value of ranking loss, the better the performance.

\begin{equation}
\begin{aligned}
RankingLoss=\frac{1}{m} \sum_{i=1}^{m} \frac{1}{|y^i||\overline{y^i}|}\cdot \#\{(\hat{y}_j,\hat{y}_k)| \\
C(x^i,\hat{y}_j)\leq{C(x^i,\hat{y}_k)},(\hat{y}_j, \hat{y}_k)\in{y^i}\times{\overline{y^i}}\}
\end{aligned}
\end{equation}

where $\overline{y^i}$ denotes the complementary set of $y^i$ in \textit{Y}.

(5) \textbf{Average Precision}: evaluates the average fraction of labels ranked above a particular label \textit{y} {$\in$} \textit{Y} which actually are in \textit{Y}. It is originally used in information retrieval (IR) systems to evaluate the document ranking performance for query retrieval. The performance is perfect when average precision equals to 1; the bigger the value of average precision, the better the performance.

\begin{equation}
\begin{aligned}
&AveragePrecision= \frac{1}{m} \sum_{i=1}^{m} \frac{1}{|y^i|} \sum_{\hat{y}\in{y^i}} \\ &\frac{\#\{y^{\prime}|rank[C(x^i,y^{\prime})]\leq{rank}[C(x^i,\hat{y})],y^{\prime}\in{y^i}\}}{rank[C(x^i,\hat{y})]}
\end{aligned}
\end{equation}

\subsection{Comparing Methods}
As we introduced above, different feature extraction methods have the huge impact on prediction performance. So we compare our framework to other methods with different schemes for extracting different level representations.

The following feature extraction schemes are adopted:

\textbf{Linear Combination}: rather than information loss during learning abstract features, directly linear combination can maintain fully original information;

\textbf{Principal Component Analysis (PCA)}: to retain key features in original data and avoid compressing data too much by PCA, at least 99\% variance is guaranteed to be retained for each type of feature;

\textbf{Autoencoder (AE)}: for extracting more representative features, we build an autoencoder by stacking two layers, and each layer is trained in unsupervised manner.

The following models are adopted for comparisons:

\textbf{Random Forest (RF)}: as a classical robust ensemble classifier, RF has capacities to eliminate the disadvantage of instability for the decision tree and cope with large feature space. In this study, the dimension of all original data is over 10,000;

\textbf{Gradient Boosting Decision Tree (GBDT)}: produces a prediction model in the form of an ensemble of weak prediction models, and optimizes a cost function over function space by iteratively choosing a function (weak hypothesis) that points in the negative gradient direction. Especially there are 7 feature sources from 3 data domains, and the target prediction task is imbalanced labeled in this paper;

\textbf{CNN}: owing to its performance of dealing with arbitrary sized data, we optimize the architecture of DeepSEA to embed heterogeneous data. Followed by logistic regression (LR) for multiple binary classifications;

\textbf{Hybrid Model}: the combination of CNN and Bi-LSTM can not only map heterogeneous data into the same feature space, also model the dependency automatically.

\subsection{Performance Assessment}
Table.2 shows the hamming loss, one error, coverage, ranking loss, and average precision of different methods.

\textbf{Experiments on several data domains.} Firstly, to evaluate the effect of knowledge sharing for prediction task, we use RF and GBDT on only drug physicochemical property (80 dimensions), only gene expression (12328 dimensions), and the linear combination of both (12408 dimensions). While the integration of two domains does not contribute to the final performance generally. For example, although the performance of RF on gene expression gets improvement by integration, the results on the only physicochemical property are still best. As well as the results of GBDT show the same tendency. It turns out that inefficient domain integration decreases the performance resulting from the feature patterns vary significantly. All results for different number of data domains are shown in Supplementary Table 1.

\textbf{Experiments on different feature extraction schemes.} Secondly, to extract different level features and explore the underlying nonlinear dependency among all heterogeneous data. PCA is adopted for selecting key features. AE and CNN are responsible for learning high level features. Then features are appended together and regarded as input of models. From these comparisons, the hybrid model with Bi-LSTM that is capable of modeling dependency among heterogeneous data leads to accurate prediction, which proves nonlinear dependency has better prospect for inference.

\textbf{Experiments on domain-specific extraction.} Eventually, in order to promote domain knowledge integration, we devised domain-specific extractor to reduce the redundancy inside heterogeneous data. Specifically, we utilize the adversarial strategy in common layers of embedding to make each feature type be classified accurately. Then, Bi-LSTM models the dependency among domain-specific features under the framework of multi-task learning. As a result, all evaluation metrics of our framework get the further improvement owing to domain-specific extractor provides prior knowledge and optimizes the searching space.


\begin{table}[!th]
    \caption{Patents of 10 Predicted Compounds}
    \label{tab:pro}
    \centering
    \begin{tabular}{lll}
        \toprule
        \textbf{BRD ID} & \textbf{PubChem CID} & \textbf{Patent ID} \\
        \midrule
        BRD-K08132273 & 2062 & US2015018301 \\
        BRD-K64341947 & 9844347 & US2016324856 \\
        BRD-K87696786 & 9926999 & US2013035335 \\
        BRD-K76304753 & 8691 & US2009227606 \\
        BRD-A26032986 & 65909 & US6406716 \\
        BRD-K38003476 & 5282493 & US2016263173 \\
        BRD-K67537649 & 9549305 & US6699879 \\
        BRD-A75552914 & 20507134 & US8168629 \\
        BRD-A78942461 & 3682 & US8716350 \\
        BRD-A35519318 & 656667 & US9517221 \\
        \bottomrule
    \end{tabular}
\end{table}

\subsection{Purpose Prediction of Purpose-unknown Small Molecule Compounds}
Having demonstrated the multi-task framework's ability to effectively extract features and to integrate data from different domains, we investigate whether the framework can identify novel purpose of compounds without a definite therapeutic property. By applying the framework to heterogeneous data of 3,749 purpose-unknown small molecule compounds in TDU, at least one ATC labels for 2,855 compounds are predicted. To further address whether these predictions are reliable, we visualize the activated and integrated features of the final layer in two dimensions using t-distributed stochastic neighbor embedding (t-SNE) for both FDA-approved drugs and purpose-unknown compounds (As shown in Fig. 6a) \cite{maaten2008visualizing}. The regions occupied by purpose-unknown compounds are covered by FDA-approved drugs with the same ATC label. For instance, 14 purpose-unknown compounds in the Example Cluster are predicted for nervous system indications. And the patents of 10 compounds are related to nervous system (as shown in Table 3 and Supplementary Table 2). We find that the purpose-unknown compounds are significantly closer to the FDA-approved drugs which have the same ATC labels than to others (Mann-Whitney test, \textit{P} Value $<$ 0.0001, Fig. 6b).

\begin{figure}[h]
  \centering
   	\includegraphics[width=0.45\textwidth]{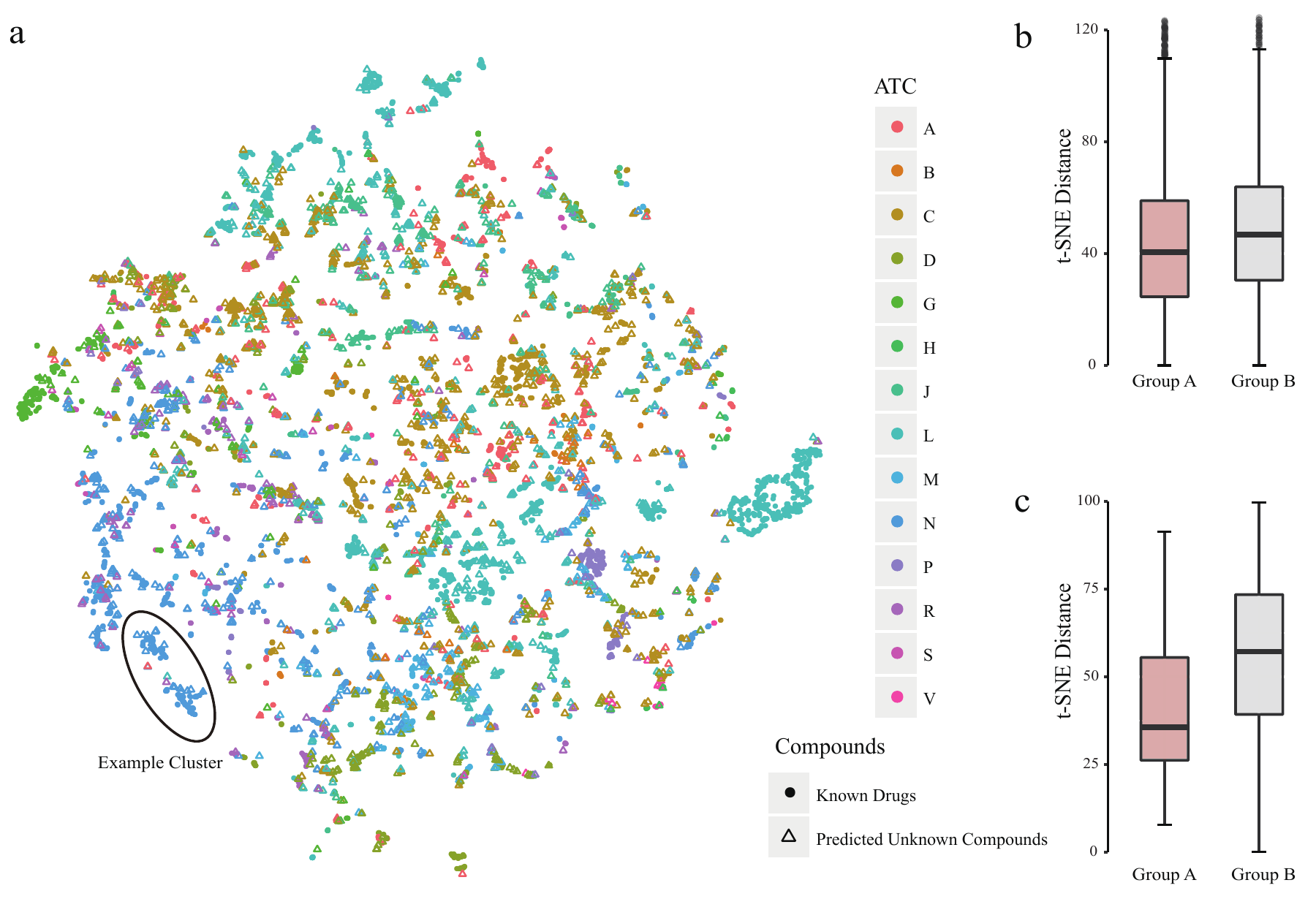}
  	\caption{Identifying novel therapeutic properties of FDA-approved and purpose-unknown compounds. (a) Comparison of purpose-unknown compounds (hollow triangle) with FDA-approved drugs (solid circle) based on t-SNE projection of features in the final layer. (b) Group A is the two-dimensional projection distance between purpose-unknown compounds with the corresponding FDA-approved drugs whose ATC is the same as predicted labels, and Group B is the distance among purpose-unknown compounds with other FDA-approved drugs whose ATC is different from predicted labels. (c) Group A is the two-dimensional projection distance between mecamylamine with nervous systems drugs, and Group B is the distance between mecamylamine with other FDA-approved drugs that are not nervous systems drugs.}
\end{figure}

\subsection{FDA-approved Drugs Repositioning}
The experimental results also show that our framework can identify novel therapeutic properties of FDA-approved drugs, namely drug repositioning. And 18 repurposed drugs have been reported or brought to the clinics, listed in a database named RepurposeDB \cite{shameer2017systematic}. For example, mecamylamine (PubChem Compound ID: 4032), the first orally available antihypertensive agent, is predicted for ¡°nervous system¡± indications. Recent researches and clinical trials demonstrated that mecamylamine is very effective for anti-depression and anti-addictive \cite{lippiello2008tc,shytle2002mecamylamine}. Moreover, in the t-SNE projection, mecamylamine is significantly closer to ¡°nervous systems¡± drugs than to others (Mann-Whitney test, \textit{P} Value $<$ 0.0001, Fig. 6c).

\section{Conclusion}
In this paper, we propose a domain-adversarial multi-task framework for joint learning representations of heterogeneous data to predict the potential purpose of small molecule compounds. Our framework can adaptively fit the data from different domains by utilizing adversarial strategy to extract domain-specific features and modeling the nonlinear dependency among heterogeneous data. Experimental results with real-world drug-informatics data prove the effectiveness of our proposed framework over competitive baselines. And as a highly extensible framework, our framework can be applied to various attributes beyond the three domains showed here. More importantly, the novel therapeutic properties of compounds we predicted stay in step with their patents, illustrating the effectiveness of our framework in the industry for screening the purpose of huge amounts of unidentified compounds in a fast and precise manner.


\bibliography{Domain-Adversarial_Multi-Task_Framework_for_Novel_Therapeutic_Property_Prediction_of_Compounds}
\bibliographystyle{aaai}
\end{document}